# Exploring the Feasibility of Deep Learning Techniques for Accurate Gender Classification from Eye Images

Basna Mohammed Salih Hasan [a], Ramadhan J. Mstafa [b,c,*]

*[a] Technical College of Duhok, Duhok Polytechnic University, Duhok, Kurdistan Region, 42001, Iraq*
*[b] Department of Computer Science, College of Science, University of Zakho, Zakho, Kurdistan Region, 42002, Iraq*
*[c] PRIME Lab, Scientific Research Center, University of Zakho, Zakho, Kurdistan Region, 42002, Iraq*

*\*Corresponding Author Email:* ramadhan.mstafa@uoz.edu.krd

**ABSTRACT**: Gender classification has emerged as a crucial aspect in various fields, including security, human-machine interaction, surveillance, and advertising. Nonetheless, the accuracy of this classification can be influenced by factors such as cosmetics and disguise. Consequently, our study is dedicated to addressing this concern by concentrating on gender classification using color images of the periocular region. The periocular region refers to the area surrounding the eye, including the eyelids, eyebrows, and the region between them. It contains valuable visual cues that can be used to extract key features for gender classification. This paper introduces a sophisticated Convolutional Neural Network (CNN) model that utilizes color image databases to evaluate the effectiveness of the periocular region for gender classification. To validate the model's performance, we conducted tests on two eye datasets, namely CVBL and (Female and Male). The recommended architecture achieved an outstanding accuracy of 99% on the previously unused CVBL dataset while attaining a commendable accuracy of 96% with a small number of learnable parameters (7,235,089) on the (Female and Male) dataset. To ascertain the effectiveness of our proposed model for gender classification using the periocular region, we evaluated its performance through an extensive range of metrics and compared it with other state-of-the-art approaches. The results unequivocally demonstrate the efficacy of our model, thereby suggesting its potential for practical application in domains such as security and surveillance.

## 1. INTRODUCTION

Gender prediction is a crucial aspect of human interaction as gender plays a significant role in social dynamics. It can be considered as a "soft biometric" feature in the field of biometrics, which refers to non-unique characteristics that can complement traditional biometrics such as fingerprints and iris scans [1]. The use of soft biometrics can enhance identification accuracy and prove useful in situations where traditional biometrics are ineffective, such as when an iris scan is unable to be obtained due to a partially closed eye. Accurate gender prediction can also play a crucial role in criminal investigations, as it can help eliminate suspects from further investigation [2]. Periocular biometrics, which refers to the area surrounding the eye, has garnered significant attention as a means of enhancing the reliability of face and iris biometric technologies. The periocular region has been widely recognized as a highly discriminative aspect of the face, as depicted in Figure 1, and has been demonstrated to be useful for independent identification. Additionally, the utilization of periocular features has been shown to improve iris recognition in situations where the intrinsic biometric information within an image is insufficient. Furthermore, research suggests that periocular characteristics can be employed for soft biometric classification [3].

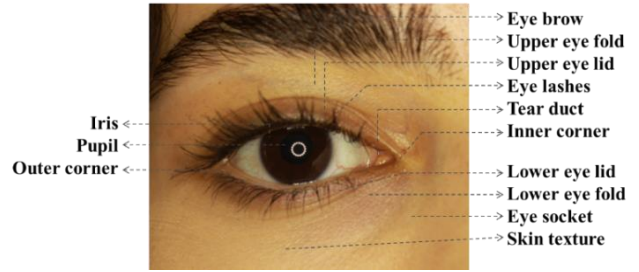

**Figure 1.** Representation of the Periocular Region of the Eye.

Traditional facial biometric systems have primarily focused on using the full face as the region of interest for identification purposes. However, these systems tend to exhibit poor performance when certain parts of the face are obscured. In response to this limitation, recent research has shifted towards utilizing the periocular region, specifically the area around the eyes, for biometric identification. The majority of these periocular algorithms holistically approach the problem, creating a region of interest that encompasses the entire eye. This comprehensive approach has the potential to include non-essential elements, such as hair or spectacles, which may negatively impact the performance of the system. Additionally, it is possible that some characteristics within the periocular region may not be equally discriminative, highlighting the need for a more targeted approach [4,5].

The periocular region plays a critical role in the soft biometric classification and matching of facial images that have undergone medical changes such as gender transition, facial surgery, and cataract surgery. This biometric attribute is particularly attractive for security and surveillance applications due to its low user involvement requirements, even when the face is partially obscured [6]. This is demonstrated in Figure 2.

The eyes can reveal information about a person's gender through distinct characteristics. Males tend to have a higher hairline and a broader forehead compared to females, who typically have bigger eyes and more arched eyebrows. Additionally, men have thicker eyebrows and a shorter gap between their eyebrows and eyes, while women have longer eyelashes and more open eyelids. Previous research has also indicated that the iris displays information about gender. These visual cues in the eyes provide a basis for determining gender in an individual [6].

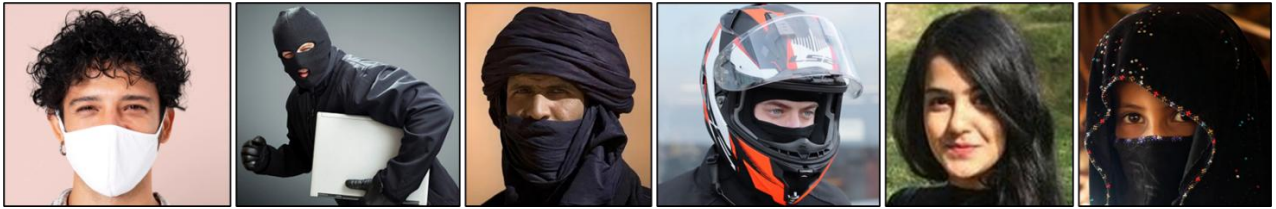

**Figure 2.** Depiction of Masked Faces [7].

Automatic gender classification from facial images is a widely researched topic in the field of computer vision and machine learning. The success of such a system is heavily dependent on the feature extraction and classification methods employed. In recent years, the availability of large face image datasets has enabled the development of advanced machine learning and deep learning techniques. Traditional machine learning approaches require careful feature extraction from the dataset to achieve high classification accuracy. However, deep learning models have revolutionized this process by automatically extracting relevant features from the raw data. This automation not only streamlines the feature extraction process but also contributes to the overall accuracy of the classification system [8].

Deep Neural Networks (DNNs) have proven to be effective in uncovering hidden and non-obvious feature sets, leading to improved classification accuracy when compared to traditional machine learning techniques. In particular, CNNs have demonstrated remarkable success in addressing the complex task of gender classification. The utilization of CNNs helps to mitigate the issue of variability in facial signals across different origins, which can impede the accuracy of feature extraction using other methods. With the advent of advanced pretrained CNN architectures, the task of image classification has become increasingly tractable. Furthermore, the scalability of CNNs is a distinct advantage, particularly when dealing with large quantities of input data, where they have been shown to consistently perform well [9].

Gender classification based on periocular images can be utilized in access control systems to enhance security in restricted areas. By incorporating gender recognition capabilities, these systems can verify the gender of an individual before allowing or denying access. This can be particularly beneficial in establishments or institutions that require gender-specific access permissions, such as gender-segregated areas or facilities following gender-based security protocols.

The rest of the paper is organized as follows: In Section 2, an extensive review of related works is provided. The CNN model is discussed in Section 3. The datasets and materials are illustrated in Section 4. The methodology of the proposed model is explained in Section 5. The experimental results of the proposed model are discussed in Section 6. Finally, the conclusion of the study is presented in Section 7.

## 2. RELATED WORKS

In 2022, Khellat-Kihel *et al.* [10] conducted a study on gender and ethnicity recognition using deep neural networks with visual attention. They proposed a deep architecture that focused on the periocular region and analyzed feature maps to extract visual saliency. The study compared different pretrained models, such as AlexNet and ResNet-50, and found that using features from earlier layers of the network improved the discrimination performance. The results demonstrated that the suggested method of using visual attention-based features on periocular regions was feasible and robust. In the same year, La Rocca *et al.* [11] presented the use of periocular data for demographic classification, specifically age and gender recognition. The authors employed a data fusion approach by combining pupil, fixation, and blink periocular features. To evaluate the system's reliability, they used transformation-based scores and classifier-based score fusion methods, including weighted sum, weighted product, and Bayesian rule. The results showed that the multi-biometric system optimized its performance and achieved an accuracy of 84.45% for age classification and 84.62% for gender classification. The fusion of these soft biometric characteristics effectively balances privacy protection and discrimination without compromising individual privacy.

In 2021, Amri *et al.* [12] conducted a study to determine the most significant facial feature for gender recognition using CNNs. They found that the eyes provided the most discriminative information, achieving an accuracy of 92% in gender identification. The study also highlighted the contributions of the mouth and nose, which achieved accuracies of 91% and 88%, respectively. These results demonstrate improved accuracy in gender classification. In another study by Cimtay *et al.* in the same year [13], the authors evaluated the performance of pre-trained CNNs for gender classification using only images of the eyes. They focused on the regions surrounding the eyes and eyebrows rather than the entire face. The study found that NasNet-Large and Xception models were the most effective for gender recognition. However, models with a larger number of parameters were more time-consuming. These findings contribute to the development of gender recognition techniques based solely on facial features.

In 2020, Abdalrady and Aly [14] proposed a method employing fusion of simple CNNs, specifically utilizing principal component analysis networks (PCANET), trained on different patch sizes. They utilized whitening PCA to reduce feature vector dimensionality. However, PCA's effectiveness in discrimination is not guaranteed. In the same year, Kumari and Seeja [15] investigated periocular biometrics using deep CNNs and transfer learning, focusing on challenging conditions like image position variation and matching periocular areas from opposite sides. VGG-19 showed high accuracy for position variation, while ResNet-18 performed well in matching periocular regions of distinct faces. They also introduced the use of reflected images for enhanced matching.

In 2019, Viedma *et al.* [16] found that the periocular area contains more gender-related information than the iris in near-infrared images. Using XGBoost and 4000 periocular characteristics, they achieved a gender classification accuracy of 89.22%. The study suggests that periocular NIR images can be used for gender classification without relying on iris data, highlighting the significance of the region around the iris.

In 2019, Kuehlkamp and Bowyer [17] studied the difficulty of identifying gender based solely on iris texture. They aimed to determine the location of gender prediction data in the periocular area or the iris stroma or both. The authors evaluated the performance of linear Support Vector Machine (SVM) and CNN in gender prediction by comparing and contrasting manual and deep features. The results of the study suggest that the periocular area contains the most significant information regarding gender. They found that the choice of kernel affects the results of SVM. The authors utilized a larger dataset for gender from the iris and applied probabilistic occlusion masking to gain a deeper understanding of the results. In the same year, Tapia *et al.* [18] conducted a study utilizing periocular iris images taken with smartphones to classify gender. Low-resolution images were enhanced using a Super-Resolution Convolutional Neural Network (SRCNN) technique. The study found that the accuracy of gender classification improved with higher image resolution, as determined by a random forest classifier. The highest classification rate was achieved for the right eye at 90.15% and for the left eye at 87.15%. The improvement was obtained by increasing the resolution of the images from 150×150 to 450×450. These results demonstrate that the SRCNN technique is effective in improving the accuracy of gender classification and aligns with current state-of-the-art results.

Finally, in a study by Manyala *et al.* in 2018 [19], investigated gender recognition using near-infrared (NIR) periocular images. They proposed two CNN-based methods that involved automated periocular area detection and extraction. The first method used deep features extracted by a pre-trained CNN and fed into an SVM for gender classification. The second method employed a CNN-based classifier trained on periocular images. Evaluation on three databases showed that the proposed methods outperformed baseline algorithms, particularly on a public dataset with imperfect images.

Existing methods for gender classification based on the periocular region have made significant progress. However, a comprehensive comparison with other state-of-the-art approaches is often lacking. To address this limitation, our study aims to evaluate the performance of our proposed model against other existing methods thoroughly. We employ a diverse set of metrics to showcase the superiority of our approach. This study underscores the potential of utilizing periocular features, particularly the regions surrounding the eyes, for accurate gender recognition. Deep learning models, such as CNNs, have demonstrated promising outcomes in extracting distinctive features from periocular images. The periocular region offers valuable cues for gender classification, often surpassing the performance of iris-based approaches. However, it is crucial to acknowledge that the accuracy of gender classification systems can vary depending on factors such as the quality of the dataset, image resolution, and the specific methodology employed.

## 3. CONVOLUTIONAL NEURAL NETWORKS

The field of deep learning falls under the umbrella of machine learning, emphasizing the utilization of deep neural networks to derive high-level abstractions from data. In recent years, there has been a notable surge in applying deep learning methods to address diverse artificial intelligence challenges, including semantic parsing, natural language processing, transfer learning, and computer vision. This growth can be attributed primarily to three factors: increased processing power in chips, a decline in the cost of computer hardware, and substantial advancements in machine learning techniques.

Deep learning algorithms exhibit a hierarchical structure, where each consecutive layer of neurons is linked to the preceding layer through electrical impulses. The input and output layers' neurons are clearly defined, while the intermediate layers are termed "hidden" layers. These networks are often expressed mathematically, using equations to represent the virtual neurons in an Artificial Neural Network (ANN). These mathematical formulations draw inspiration from the structural biology of the brain [20]. In the accompanying illustration, we present a depiction of a single neuron within a deep learning network, featuring the input, the neuron's weight, and the bias (denoted as "b"), as shown in Figure 3. The

neuron's output is computed based on these parameters in the following manner:

$$Y = \sigma(w \cdot x + b) \quad (1)$$

Where σ represents an activation function.

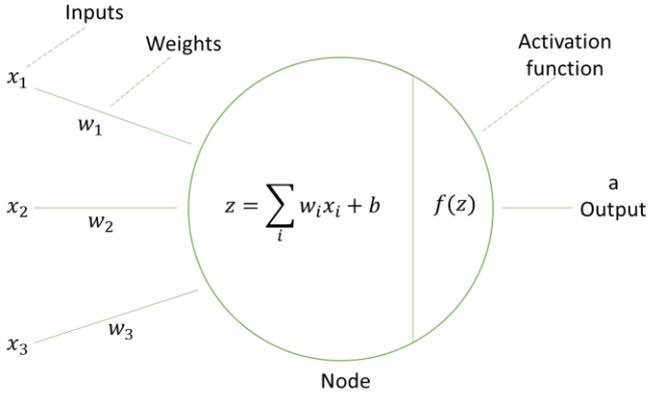

**Figure 3.** Components of a Layered Neural Network [21].

CNN represents a pervasive deep learning methodology, constituting the most prevalent category of feed-forward deep neural networks. Their widespread adoption is particularly evident in the realm of computer vision, notably in applications such as medical image analysis. A conventional CNN comprises two fundamental types of layers in its early stages: convolutional layers and pooling layers. The training of CNN layers follows a methodical process.

In the initial stages, convolutional layers of a CNN take an image as input and generate feature maps as output. Each feature map is crafted through a convolution process involving filters, and the weighted sum of convolutions is subjected to non-linear functions such as ELU. Diverse feature maps arise by employing distinct sets of filters, although a consistent set of filters is shared across all neurons within a singular feature map [22]. The weighted sum of the convolutions is represented by the following equation and represented as *k* feature map.

$$y^k = \sigma(\Sigma_m w_m^k * x_m + b_k) \quad (6)$$

Where the feature maps from the inputs are added together, the asterisk (*) stands for the convolution operator and the filters.

A pooling layer in CNNs serves the purpose of reducing the spatial dimensions of the feature maps generated by the convolutional layer. This reduction in the spatial size not only reduces the number of network parameters but also reduces the computational load, making the network more efficient. Pooling layers work by applying an operation, typically the max operation, independently on each depth slice of its input. This operation is performed with a stride that is similar to the stride used in the convolutional layer's filters. The CNNs are neural networks that are specifically designed to process image data. These networks are composed of multiple layers, as shown in Figure 4, and are capable of identifying patterns in the input image with minimum preprocessing. These models have a high learning capacity and can be trained from scratch or fine-tuned using the Transfer Learning technique. Pooling layers play a crucial role in the design of CNNs and help to make the network more efficient and effective in recognizing patterns in the input image [23–25].

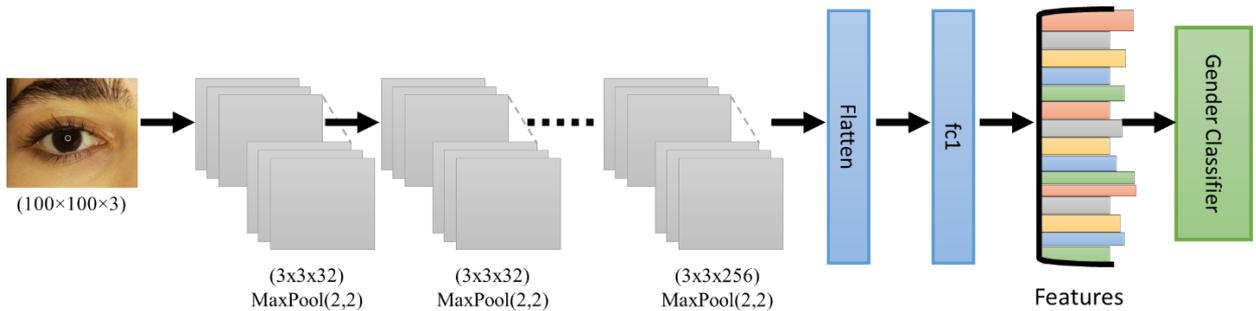

**Figure 4.** Diagram of a CNN Architecture for Classification [24].

**3.1 Exponential Linear Units (ELUs) Function**

The Exponential Linear Unit (ELU) is a variant of the popular Rectified Linear Unit (ReLU) activation function, which has become a staple in deep learning architectures. ELU differs from ReLU in its handling of negative inputs. While ReLU simply sets all negative values to zero, ELU uses a logarithmic curve to model the negative half of the activation function. The parameter α, referred to as the "alpha constant," determines the smoothness of the curve for negative inputs, as demonstrated in Figure 5. The impact of α on the negative half of the ELU function can be explored through the interactive equation provided in the literature [26,27] as shown below:

$$ELU(x) = \begin{cases} x & if\ x >= 0 \\ \alpha(e^x - 1) & if\ x < 0 \end{cases} \quad (2)$$

Two of the most widely used activation functions in artificial neural networks are the ELU and the ReLU. These functions are crucial in determining a network's ability to learn complex patterns and representations. In this context, it is important to understand both the advantages and limitations of using these activation functions. The mathematical representation of ELU can be described as follows:

$$f'(x) = \begin{cases} 1 & \text{for } x \geqslant 0 \\ f(x) + \alpha & \text{for } x < 0 \end{cases} \quad (3)$$

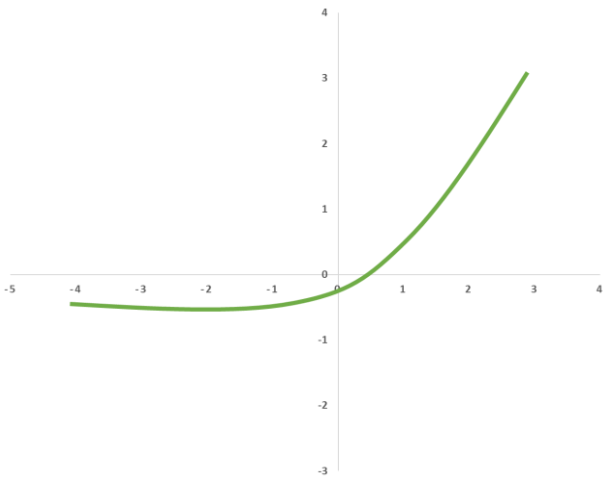

**Figure 5:** Graphical Representation of the ELU Activation Function.

The use of the ELU activation function in neural networks has been shown to result in faster convergence during the training phase compared to the ReLU. Despite the added computational cost of ELU, its non-zero gradient for negative inputs allows for better preservation of the intermediate activations, leading to improved overall accuracy. However, it should be noted that ELU is computationally more expensive than ReLU during the testing phase. Figure 6 illustrates the derivative of the ELU activation function [28].

The Exponential Linear Unit (ELU) exhibits robust capability in handling negative values, preventing dead neurons from occurring in the network, and ensuring sustained activity of neurons even with negative inputs. This characteristic contributes to the overall robustness of the network. Moreover, ELU's flexibility in capturing both positive and negative values encourages the model to explore a broader feature space. This adaptability proves particularly advantageous in complex tasks, where various features play a role in the learning process, enabling the model to effectively learn and represent intricate patterns in the data.

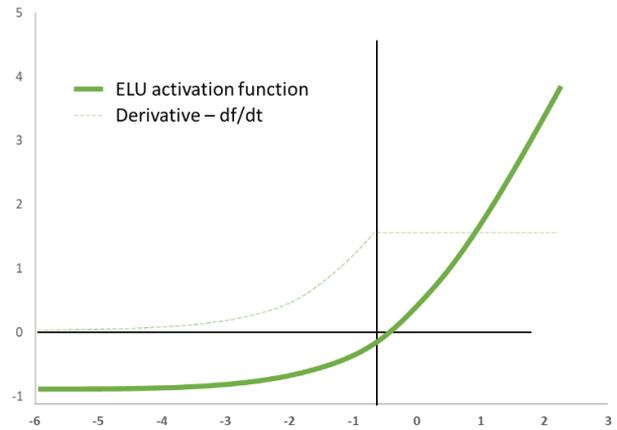

**Figure 6.** Graphical Representation of the ELU Activation Function and its Derivative.

### 3.2 Sigmoid / Logistic Activation Function

The sigmoid function, specifically the sigmoid activation function, is commonly utilized in models that require the prediction of a probability output. This is due to the fact that probabilities are defined as values between 0 and 1. The mapping is such that, given a real number x as input, the output y will approach 1.0 as x approaches positive infinity, and y will approach 0.0 as x approaches negative infinity. This relationship can be visualized as shown in Figure 7 [29].

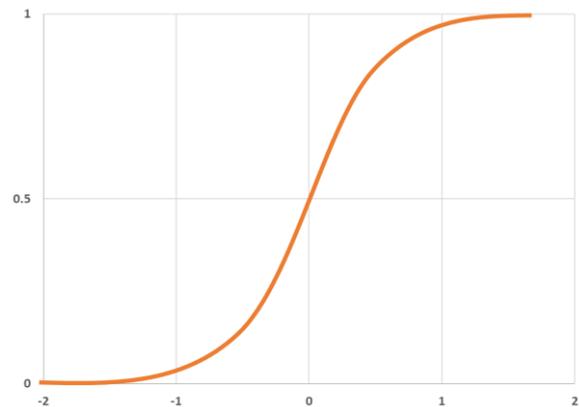

**Figure 7.** Graphical Representation of the Sigmoid/Logistic Activation Function [30].

Mathematically, it can be represented as:

$$f(x) = \frac{1}{1+e^{-x}} \quad (4)$$

The sigmoid function is often used in models that need the prediction of a probability output. It satisfies this requirement, as it maps any input value to a value between 0 and 1. Additionally, the sigmoid function is differentiable and has a smooth gradient, which helps to prevent output values from experiencing abrupt jumps [31]. The smooth, S-shaped curve of the sigmoid activation function is depicted in Figure 8. The derivative of the sigmoid function can be expressed as follows:

$$f'(x) = sigmoid(x)*(1 - sigmoid(x)) \quad (5)$$

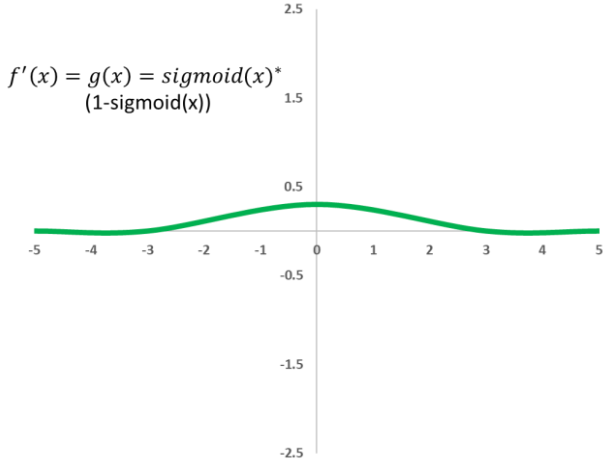

**Figure 8.** Graphical Representation of the Derivative Sigmoid Activation Function [30].

### 3.3 Stochastic Gradient Descent (SGD)

SGD is a popular optimization technique for maximizing differentiable or sub-differentiable objective functions. It approximates gradient descent by randomly selecting subsets of data to estimate the gradient. This approach speeds up iterations but sacrifices convergence rate. SGD is beneficial for high-dimensional problems with heavy computational loads [32]. The θ parameters of the objective function J(θ) are updated using the regular gradient descent technique in the following way:

$$\theta = \theta - \alpha \nabla_\theta E[J(\theta)] \quad (7)$$

The cost and gradient can be estimated by evaluating them on the entire training set, providing an approximation of the expected value. In contrast, SGD calculates the gradient using a small number of randomly selected training samples, eliminating the expectation from the update. See Figure 9 for an illustration of this process. This latest revision is provided as follows:

$$\theta = \theta - \alpha \nabla_\theta J(\theta; x^{(i)}, y^{(i)}) \quad (8)$$

with a pair $(x^{(i)}, y^{(i)})$ from the training set.

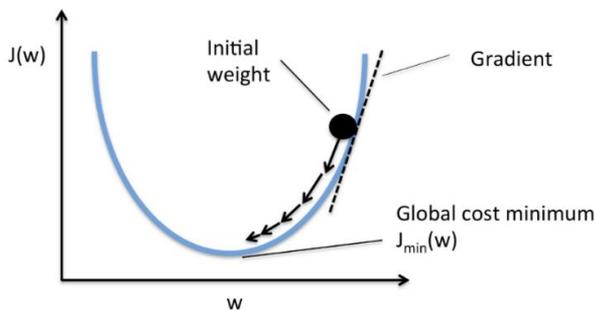

**Figure 9.** Graphical Representation of the Stochastic Gradient Descent [33].

## 4. MATERIAL

The utilization of various face datasets has been a common practice in the field of gender classification. These datasets have proven to be effective in capturing facial features and facilitating accurate classification results. However, most of these datasets provide images of the entire face, which can lead to increased computational resources and added study effort in the form of segmenting the eyes for analysis. In this study, we aim to address this issue by utilizing two datasets, namely the CVBL dataset and the Female and Male datasets. These datasets have been chosen for their diverse representation of facial features, which will enable us to analyze and compare the performance of different gender classification models. The use of these datasets will allow us to evaluate the efficacy of our proposed approach in terms of accuracy and computational efficiency, while also contributing to the existing literature on gender classification.

### 4.1 CVBL Dataset

The CVBL IRIS gender classification database [34] is a seminal study in the field of gender classification based on eye images. The dataset was collected from 720 university students, with a balanced distribution of 370 female and 350 male participants. To ensure the reliability and consistency of the data, more than 6 images were captured from each student's left and right eyes, with three images from each eye. The images were captured under identical conditions using a color camera. The resulting dataset consists of 4320 images, providing a comprehensive representation of the male and female populations. As demonstrated in Figure 10, the dataset includes a range of eye images for both male and female participants. The detailed description of the dataset can be found in Table 1.

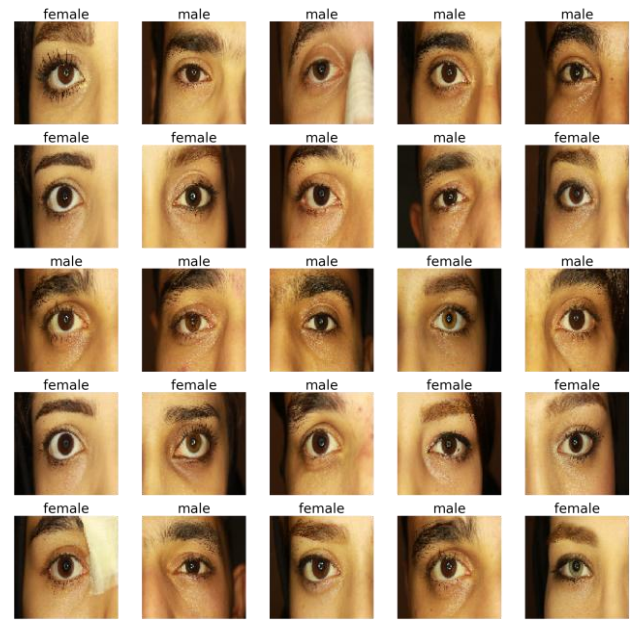

**Figure 10.** Samples of Male and Female Eye Images in the CVBL Dataset.

## 4.2 Female and Male Dataset

Additionally, this study utilized the "Female and Male" dataset, as referenced in [35]. This dataset specifically comprises eye images that have been extracted from full face images. It is noteworthy that these images often include complete or partial eyebrows. The dataset consists of 5202 images of female eyes and 6323 images of male eyes, providing a comprehensive representation of the female and male populations. As depicted in Figure 11, this dataset includes examples of eye images for both female and male participants.

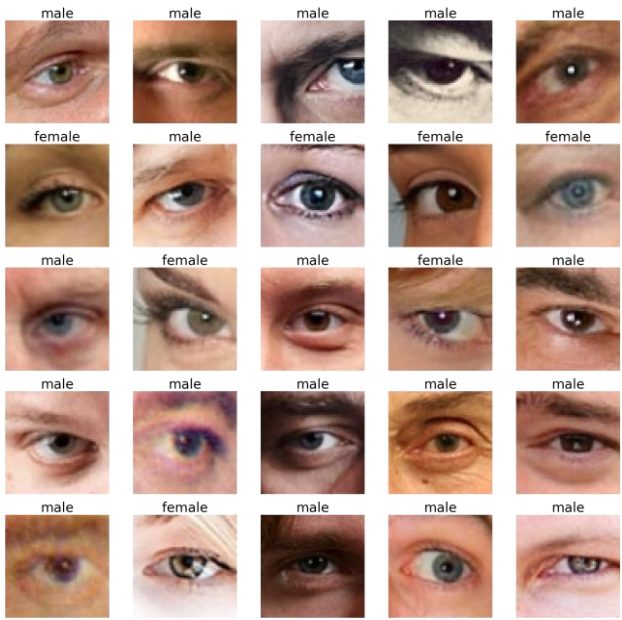

**Figure 11.** Samples of Male and Female Eye Images in Female and Male Dataset.

## 5. THE PROPOSED CNN MODEL

The proposed deep learning architecture consists of a multi-layered CNN designed specifically for periocular gender recognition. The architecture includes 10 convolutional layers, which are responsible for feature extraction, and five fully connected layers, which are responsible for classification. The input layer, which is the first layer, accepts an image of 100x100 pixels with three RGB channels. The convolutional layer, which is the second layer, has a convolution window size of 3x3 pixels and the same padding. The nonlinear activation function used in the architecture is the ELU, which follows the convolutional layer. The normalization and max pooling layers with a window size of 2x2 are then applied. All the subsequent convolutional layers are also followed by an ELU activation function, a batch normalization layer, and either a max pooling layer or a dropout layer, depending on the requirement.

A fully connected layer comprising 4096 neurons, incorporating the ELU activation function, is succeeded by a dropout layer designed to mitigate overfitting. Subsequently, two successive fully connected layers are implemented, featuring 1024 and 128 neurons, each employing ELU activation functions. The final fully connected layer is equipped with 2 neurons, responsible for classifying outcomes into 2 distinct categories for periocular gender recognition. A sigmoid layer is subsequently employed to ascertain class membership, predicting whether the input image pertains to the female or male class. The structure of the suggested CNN is depicted in figures 12, 13, and 14, and the details of the suggested structure are provided in Table 2.

**Table 1.** Description of CVBL and "Female and Male" Datasets

| Details | Description CVBL | Female and Male |
|---|---|---|
| Total Number of Images | 4320 images | 11525 images |
| Image Capturing Device | Canon D550 with 18-55 lens. | Collected from the site https://ruskino.ru/ |
| Number of male images | 350 for men | 6323 for men |
| Number of female images | 370 for women | 5202 form women |
| Number of images per eye | 3 image per eye | 1 |
| Resolution of image | 5184×3456 | Different sizes |
| Number of subjects | 720 | - |
| Format | Jpg | Jpg |

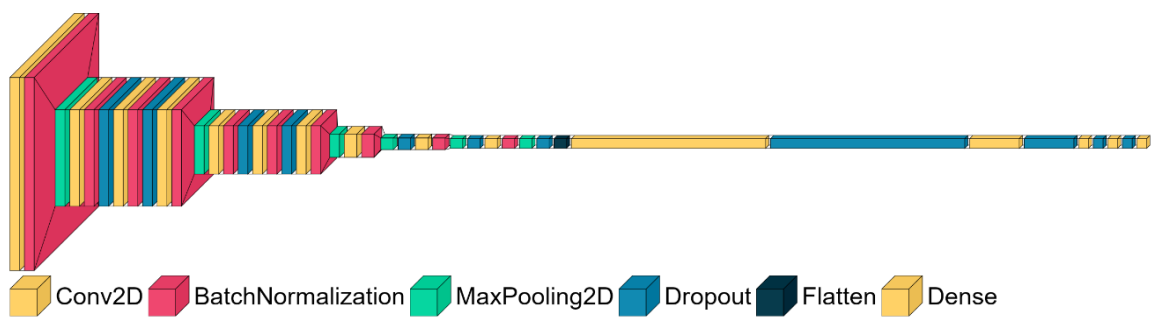

**Figure 12.** The Proposed CNN Architecture Model.

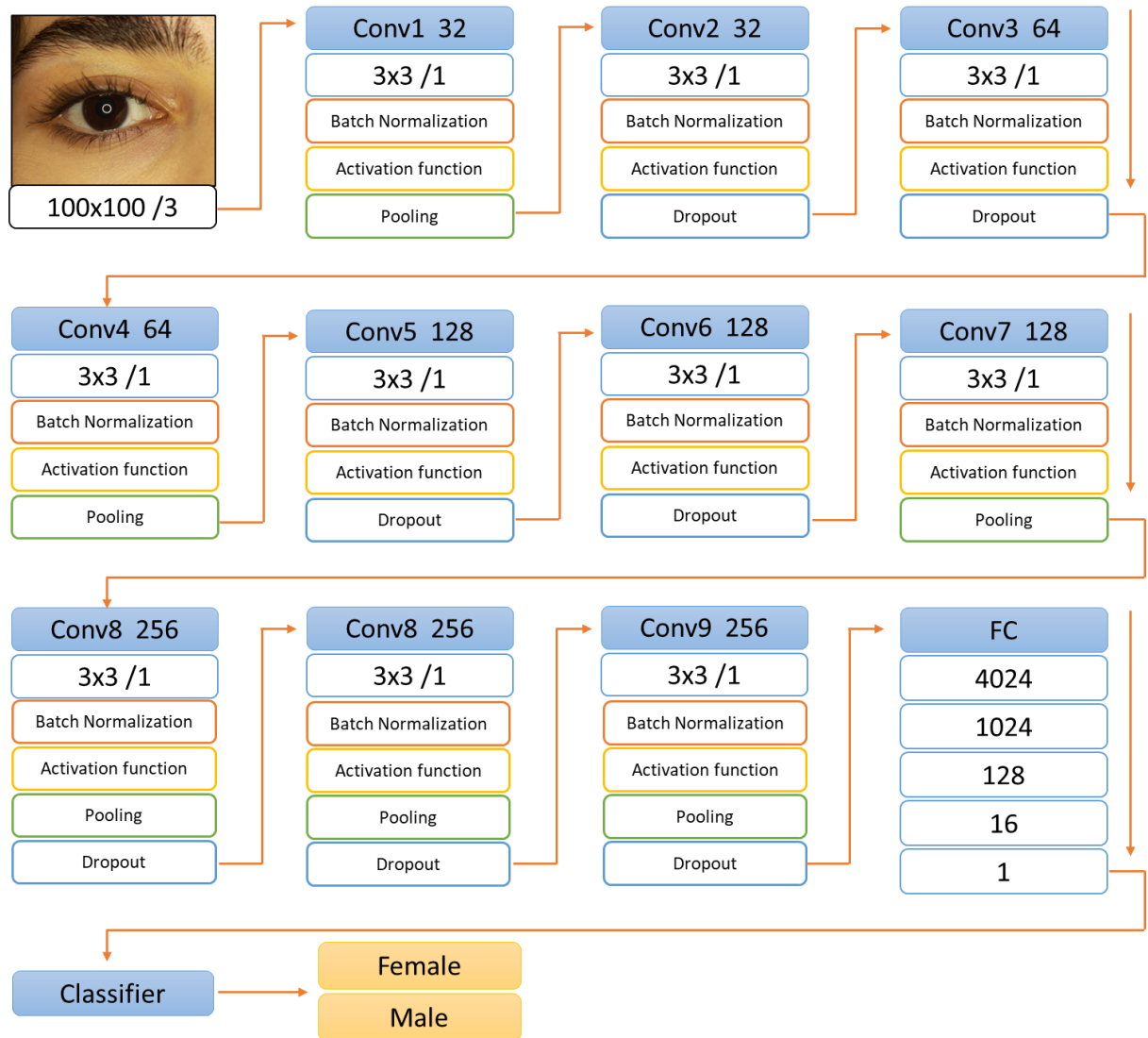

**Figure 13.** The Framework of the Proposed CNN Model.

**Table 2.** The Detailed Parameters of the Proposed CNN Model.

| Type of Layer | Configurations |
| --- | --- |
| Convolutional Layer | Convolutional (3,3),3, padding =same, stride=1 |
| Batch Normalization | Mini-batch size 64 |
| Activation layer | ELU |
| Pooling layer | Max Pooling, Pooling size=2,2 |
| Dropout layer | Dropout probability = 0.15 |
| Fully-Connected | Number of outputs (classes)=1 |
| Classifier | Sigmoid |
| Total parameters | 7,235,089 |
| Trainable parameters | 7,232,401 |
| Non-trainable parameters | 2,688 |
| Optimizer | SGD |
| Loss | Binary Cross-Entropy |
| Number of epochs | 200 |
| Batch size | 128 |

## 6. RESULTS AND DISCUSSION

The proposed architecture in this study was developed utilizing the software package Python. It was designed to be GPU-specific, and the experiments were conducted on a computer system equipped with an 11th Generation Intel(R) Core (TM) i7-1185G7 processor with a clock speed of 3.00 GHz and 16.0 GB of RAM. To validate the performance of the proposed architecture, two eye datasets were employed. The datasets were divided into 70% for training and 30% for validation and testing purposes. The training process was carried out using the Keras library in a Python environment.

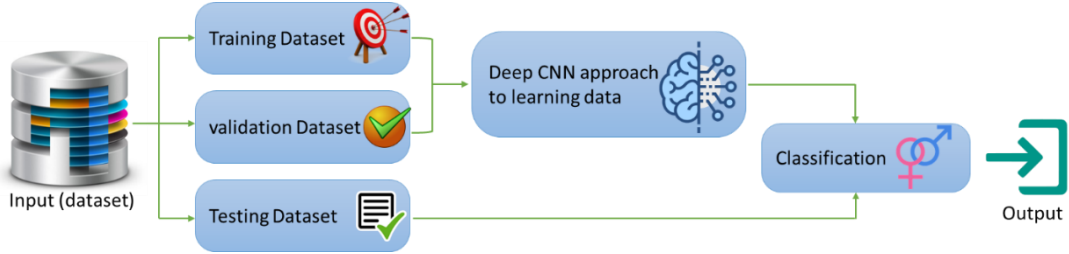

**Figure 14.** A Description of the Division of Data into Training, Validation, and Testing Sets.

### 6.1 Trained CNN Model using CVBL Dataset

The proposed model in this study was implemented on the CVBL dataset after undergoing normalization. The dataset was then divided into three subsets. The first subset, containing 3114 images, was utilized for training and constituted 70% of the total dataset. The second and third subsets, each containing 554 images, were utilized for testing and verification, respectively, and constituted 30% of the total dataset each. The approach used in this study is depicted in Figure 14. To evaluate the final testing accuracy of the proposed architecture, the median testing accuracy obtained was 99%. The gender feature representations obtained through the proposed model were used for classification and prediction with CNNs. The training procedure employed the Stochastic Gradient Descent with Momentum (SGDM) learning algorithm. To align with the input layer size of pretrained CNN models, the input image size was set to 100×100×3. A higher momentum value of 0.8 was utilized to expedite the training process. Table 3 presents the classification report for the CVBL dataset, which was used for testing purposes.

**Table 3.** Classification Report on the Test CVBL Dataset.

|  | Precision | Recall | F1-score | Support |
|---|---|---|---|---|
| **Male** | 1.00 | 0.98 | 0.99 | 327 |
| **Female** | 0.98 | 1.00 | 0.99 | 227 |
| **Accuracy** |  |  | **0.99** | 554 |
| **Macro Avg** | 0.99 | 0.99 | 0.99 | 554 |
| **Weighted Avg** | 0.99 | 0.99 | 0.99 | 554 |

Figure 15 illustrates the confusion matrix for the proposed gender classification on the CVBL dataset. This graphical representation provides an evaluation of the classifier's performance by comparing the predicted class labels with the actual class labels of the test data, offering an overall view of its accuracy.

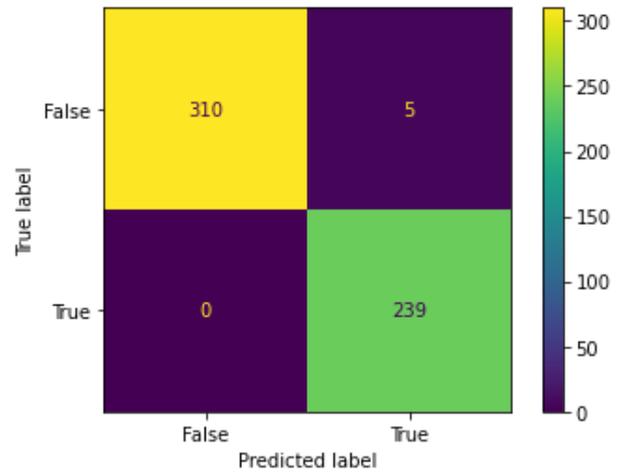

**Figure 15.** The Confusion Matrix of the Proposed Model for the CVBL Dataset.

As depicted in Figures 16 and 17, the accuracy and loss curves are presented below the model figure. These curves provide an in-depth analysis of the performance of the model during the training process. The accuracy curve indicates the rate at which the model correctly classifies the data points, while the loss curve represents the difference between the predicted and actual values.

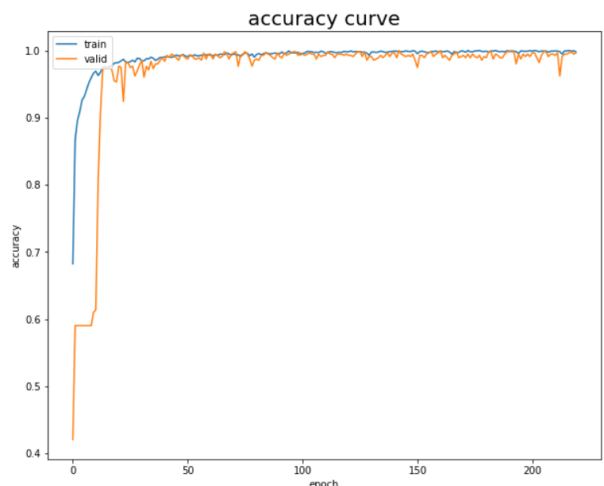

**Figure 16.** The Accuracy Curve of the Model.

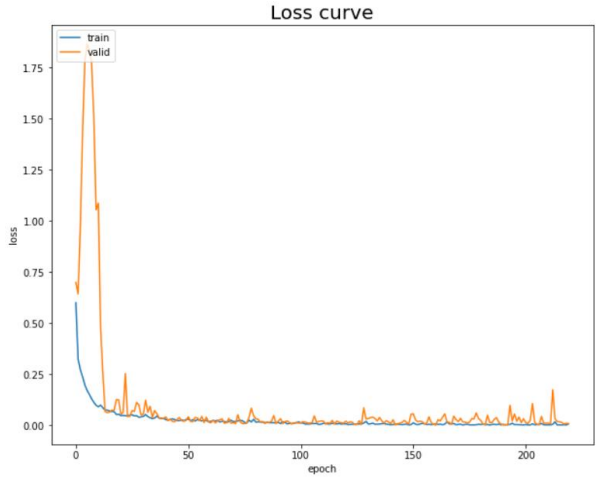

**Figure 17.** The Loss Curve of the Model.

### 6.2 Trained CNN Model using (Female and Male) Dataset

In this study, the proposed approach was applied to the (Female and Male) dataset, which consists of 6323 images of male eyes and 5202 images of female eyes. To balance the dataset, under-sampling was performed by randomly deleting 1000 rows from the majority class (male images) to match the number of images in the minority class (female images). As a result, the number of male images was reduced to 5323. The dataset was then divided into three subsets. The first subset, containing 7367 images, was utilized for training and constituted 70% of the total dataset. The second and third subsets, each containing 1579 images, were utilized for testing and verification, respectively, and constituted 30% of the total dataset each. To evaluate the final testing accuracy of the proposed architecture, the median testing accuracy obtained was 96%. Furthermore, a comparison of the proposed architecture with related works was conducted using the same dataset. The classification report for the results on the test Female and Male dataset is presented in Table 4. The study also utilized several pretrained deep models, including InceptionV3, InceptionResnetV2, Xception, and NASNetLarge, for training. The results obtained from these models were as follows: NASNetLarge and InceptionResnetV2 models achieved 95% accuracy, while InceptionV3 and Xception models achieved 96% accuracy.

**Table 4.** Classification Report on the Test Female and Male Datasets.

|  | Precision | Recall | F1-score | Support |
|---|---|---|---|---|
| **Male** | 0.96 | 0.98 | 0.97 | 945 |
| **Female** | 0.97 | 0.95 | 0.96 | 784 |
| **Accuracy** |  |  | **0.96** | 1729 |
| **Macro Avg** | 0.96 | 0.96 | 0.96 | 1729 |
| **Weighted Avg** | 0.96 | 0.96 | 0.96 | 1729 |

The confusion matrix of the proposed gender classification on the "Female and Male" dataset is presented in Figure 18. The confusion matrix provides a visual representation of the performance of the proposed architecture in terms of its accuracy and precision in classifying the images into male and female categories. The matrix presents a breakdown of the model's predictions, categorizing them into true positive (TP), false positive (FP), false negative (FN), and true negative (TN) outcomes.

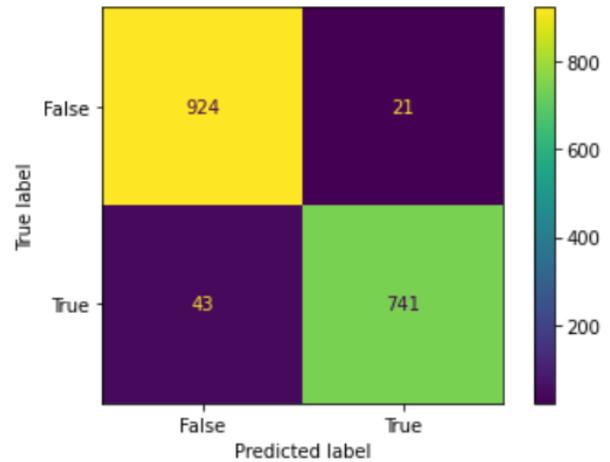

**Figure 18.** The Confusion Matrix of the Proposed Model for the Female and Male Dataset.

### 6.3 Comparison with Related Works

In the study presented by CIMTAY [13], the gender classification was performed on eye images from the (Female and Male) dataset using pre-trained CNNs, including InceptionV3, InceptionResnetV2, Xception, and NASNetLarge. The evaluation results showed that the InceptionResnetV2 and NASNetLarge models achieved 95% accuracy, while the highest accuracy was achieved by the InceptionV3 and Xception models with 96% accuracy, and with the number of parameters being 23,851,784 and 22,910,480, respectively. Our proposed model was also applied to the same dataset and achieved an accuracy of 96%, with a lower number of parameters (7,235,089).

This study demonstrates the reduction of gender classification from face images to gender classification from eye images, narrowing the region of interest and data set. Our proposed network was evaluated on the same classification problem by using the trained dataset and the results were compared with relevant works in the field, which are summarized in Table 5.

**Table 5.** Evaluation of the Proposed Model with other Existing Methods in Terms Accuracy.

| Ref. | Year | Datasets | Techniques | Area | Accuracy |
|---|---|---|---|---|---|
| Manyala [19] | 2018 | In-house, The MBGC portal and the IIITD multispectral periocular database | CNN and SVM | Periocular | 81%, 84%, 94%. |
| Tapia [18] | 2019 | Selfie database with an iPhone x | SRCNNS, random forest classifier | Periocular iris | 90.15% for the right and 87.15% for the left eye |
| Kuehlkamp [17] | 2019 | GFI, GFI-C | CNN and SVM | Periocular and iris | 80% |
| Viedma [16] | 2019 | GFI-UND, UTIRIS, cross-eyed | Xgboost, SVM | Periocular | 89% |
| Kumari [15] | 2020 | UBIPr, | VGG 19 | Periocular | 96% |
| Abdalrady [14] | 2020 | Gallagher | PCANet | Face | 89% |
| Amri [12] | 2021 | Utkface | CNN | Face | 92% for eyes, 91% for mouth and 89% for nose |
| CIMTAY [13] | 2021 | Female and Male | Xception | Eye | 96% |
| La Rocca [11] | 2022 | GANT | Fusion strategy | Periocular | 84.62% |
| **Proposed Model** | **2023** | **CVBL Female and Male** | **Deep CNN** | **Periocular** | **99% 96%** |

## 7. CONCLUSION

The field of periocular biometrics has experienced rapid advancements and has now reached a level of efficacy that competes with traditional iris and facial recognition techniques. In this study, the advancement of periocular biometrics has been explored and compared with gender recognition methods. The proposed deep CNN model has been applied to two eye datasets, including the CVBL dataset, which consists of high-resolution images. The median testing accuracy obtained from the CVBL dataset was 99%. For gender classification, the proposed CNN was trained using the SGDM learning algorithm. Additionally, the proposed architecture was compared to related works using a (Female and Male) dataset, which showed a median testing accuracy of 96%, similar to the best pre-trained models (InceptionV3 and Xception). The results of this study demonstrate the effectiveness of the proposed model in accurately determining gender from periocular images using the proposed CNN model. This method has potential applications in the fields of biometrics and security, where gender identification is a crucial factor.

**DATA AVAILABILITY:** All datasets used in this work are publicly available. The dataset CVBL IRIS gender classification database [34] is available at (https://ieee-dataport.org/documents/iris-super-resolution-dataset).
Also, the dataset Female and Male, as referenced in [35] is available at (https://www.kaggle.com/datasets/pavelbiz/eyes-rtte).

## REFERENCES

[1] Agbo-Ajala, O.; Viriri, S. *Deep Learning Approach for Facial Age Classification: A Survey of the State-of-the-Art*; Springer Netherlands, 2021; Vol. 54. https://doi.org/10.1007/s10462-020-09855-0.

[2] Vatsa, M., Singh, R., Majumdar, A. *Deep Learning in Biometrics*; 2018. https://doi.org/10.1201/b22524.

[3] Boyd, K.; Turbert, D. Eye Anatomy: Parts of the Eye and How We See - American Academy of Ophthalmology. *Am. Acad. Opthalmology website* 2021, 1–4.

[4] Rajan, B. K.; Anto, N.; Jose, S. Fusion of Iris & Fingerprint Biometrics for Gender Classification Using Neural Network. *2nd Int. Conf. Curr. Trends Eng. Technol. ICCTET 2014* 2014, 216–221. https://doi.org/10.1109/ICCTET.2014.6966290.

[5] Singh, M.; Nagpal, S.; Vatsa, M.; Singh, R.; Noore, A.; Majumdar, A. Gender and Ethnicity Classification of Iris Images Using Deep Class-Encoder. *IEEE Int. Jt. Conf. Biometrics, IJCB 2017* 2018, *2018-Janua*, 666–673. https://doi.org/10.1109/BTAS.2017.8272755.

[6] Rattani, A.; Reddy, N.; Derakhshani, R. Convolutional Neural Networks for Gender Prediction from Smartphone-Based Ocular Images. *IET Biometrics* 2018, *7* (5), 423–430. https://doi.org/10.1049/iet-bmt.2017.0171.

[7] Kumari, P.; Seeja, K. R. Periocular Biometrics: A Survey. *J. King Saud Univ. - Comput. Inf. Sci.* 2022, *34* (4), 1086–1097. https://doi.org/10.1016/J.JKSUCI.2019.06.003.

[8] Alghaili, M.; Li, Z.; Ali, H. A. R. Deep Feature Learning for Gender Classification with Covered/Camouflaged Faces. *IET Image Process.* 2020, *14* (15), 3957–3964. https://doi.org/10.1049/iet-ipr.2020.0199.

[9] Khan, H. K.; Shah, A. S.; Khan, M. A. Critical Evaluation of Frontal Image-Based Gender Classification Techniques. *Int. J. Signal Process. Image Process. Pattern Recognit.* 2016, *9* (11), 283–294. https://doi.org/10.14257/ijsip.2016.9.10.27.

[10] Khellat-Kihel, S.; Muhammad, J.; Sun, Z.; Tistarelli, M. Gender and Ethnicity Recognition Based on Visual Attention-Driven Deep Architectures. *J. Vis. Commun.*

*Image Represent.* 2022, *88*, 103627. https://doi.org/https://doi.org/10.1016/j.jvcir.2022.103627.

[11] Rocca, D. La; Bisogni, C.; Cascone, L.; Narducci, F. Periocular Data Fusion for Age and Gender Classification. *J. Imaging 2022, Vol. 8, Page 307* 2022, *8* (11), 307. https://doi.org/10.3390/JIMAGING8110307.

[12] Amri, R.; Gazdar, A.; Barhoumi, W. A Comparative Study on the Importance of Each Face Part in Facial Gender Recognition via Convolutional Neural Networks. In *2021 IEEE/ACS 18th International Conference on Computer Systems and Applications (AICCSA)*; 2021; pp 1–8. https://doi.org/10.1109/AICCSA53542.2021.9686825.

[13] CIMTAY, Y.; YILMAZ, G. N. Gender Classification from Eye Images by Using Pretrained Convolutional Neural Networks. *Eurasia Proc. Sci. Technol. Eng. Math.* 2021, *14* (March 2022), 39–44. https://doi.org/10.55549/epstem.1050171.

[14] Abdalrady, N. A.; Aly, S. Fusion of Multiple Simple Convolutional Neural Networks for Gender Classification. In *2020 International Conference on Innovative Trends in Communication and Computer Engineering (ITCE)*; 2020; pp 251–256. https://doi.org/10.1109/ITCE48509.2020.9047798.

[15] Kumari, P.; Seeja, K. R. Periocular Biometrics for Non-Ideal Images: With off-the-Shelf Deep CNN & Transfer Learning Approach. *Procedia Comput. Sci.* 2020, *167*, 344–352. https://doi.org/https://doi.org/10.1016/j.procs.2020.03.234.

[16] Viedma, I.; Tapia, J.; Iturriaga, A.; Busch, C. Relevant Features for Gender Classification in NIR Periocular Images. *IET Biometrics* 2019, *8* (5), 340–350. https://doi.org/10.1049/IET-BMT.2018.5233.

[17] Kuehlkamp, A.; Bowyer, K. Predicting Gender from Iris Texture May Be Harder than It Seems. *Proc. - 2019 IEEE Winter Conf. Appl. Comput. Vision, WACV 2019* 2019, 904–912. https://doi.org/10.1109/WACV.2019.00101.

[18] Tapia, J.; Arellano, C.; Viedma, I. Sex-Classification from Cellphones Periocular Iris Images. *Adv. Comput. Vis. Pattern Recognit.* 2019, 227–242. https://doi.org/10.1007/978-3-030-26972-2_11.

[19] Manyala, A.; Cholakkal, H.; Anand, V.; Kanhangad, V.; Rajan, D. CNN-Based Gender Classification in near-Infrared Periocular Images. *Pattern Anal. Appl.* 2019, *22* (4), 1493–1504. https://doi.org/10.1007/s10044-018-0722-3.

[20] Reddy, N.; Rattani, A.; Derakhshani, R. Generalizable Deep Features for Ocular Biometrics. *Image Vis. Comput.* 2020, *103*, 103996. https://doi.org/10.1016/J.IMAVIS.2020.103996.

[21] Khalifa, N. E. M.; Taha, M. H. N.; Hassanien, A. E.; Mohamed, H. N. E. T. Deep Iris: Deep Learning for Gender Classification through Iris Patterns. *Acta Inform. Medica* 2019, *27* (2), 96–102. https://doi.org/10.5455/aim.2019.27.96-102.

[22] Shin, H. C.; Roth, H. R.; Gao, M.; Lu, L.; Xu, Z.; Nogues, I.; Yao, J.; Mollura, D.; Summers, R. M. Deep Convolutional Neural Networks for Computer-Aided Detection: CNN Architectures, Dataset Characteristics and Transfer Learning. *IEEE Trans. Med. Imaging* 2016, *35* (5), 1285–1298. https://doi.org/10.1109/TMI.2016.2528162.

[23] Aloysius, N.; Geetha, M. A Review on Deep Convolutional Neural Networks. *Proc. 2017 IEEE Int. Conf. Commun. Signal Process. ICCSP 2017* 2018, *2018-Janua*, 588–592. https://doi.org/10.1109/ICCSP.2017.8286426.

[24] Albelwi, S.; Mahmood, A. A Framework for Designing the Architectures of Deep Convolutional Neural Networks. *Entropy 2017, Vol. 19, Page 242* 2017, *19* (6), 242. https://doi.org/10.3390/E19060242.

[25] Kim, Y.; Moon, T. Human Detection and Activity Classification Based on Micro-Doppler Signatures Using Deep Convolutional Neural Networks. *IEEE Geosci. Remote Sens. Lett.* 2016, *13* (1), 8–12. https://doi.org/10.1109/LGRS.2015.2491329.

[26] Trottier, L.; Gigure, P.; Chaib-Draa, B. Parametric Exponential Linear Unit for Deep Convolutional Neural Networks. *Proc. - 16th IEEE Int. Conf. Mach. Learn. Appl. ICMLA 2017* 2017, *2017-Decem*, 207–214. https://doi.org/10.1109/ICMLA.2017.00038.

[27] Clevert, D. A.; Unterthiner, T.; Hochreiter, S. Fast and Accurate Deep Network Learning by Exponential Linear Units (ELUs). *4th Int. Conf. Learn. Represent. ICLR 2016 - Conf. Track Proc.* 2015. https://doi.org/10.48550/arxiv.1511.07289.

[28] Qiumei, Z.; Dan, T.; Fenghua, W. Improved Convolutional Neural Network Based on Fast Exponentially Linear Unit Activation Function. *IEEE Access* 2019, *7*, 151359–151367. https://doi.org/10.1109/ACCESS.2019.2948112.

[29] Gecynalda, G. S.; Ludermir, T. B.; Lima, L. M. M. R. Comparison of New Activation Functions in Neural Network for Forecasting Financial Time Series. *Neural Comput. Appl.* 2011, *20* (3), 417–439. https://doi.org/10.1007/S00521-010-0407-3/METRICS.

[30] Rasamoelina, A. D.; Adjailia, F.; Sincak, P. A Review of Activation Function for Artificial Neural Network. *SAMI 2020 - IEEE 18th World Symp. Appl. Mach. Intell. Informatics, Proc.* 2020, 281–286. https://doi.org/10.1109/SAMI48414.2020.9108717.

[31] Datta, L. A Survey on Activation Functions and Their Relation with Xavier and He Normal Initialization. 2020. https://doi.org/10.48550/arxiv.2004.06632.

[32] Kuzborskij, I.; Lampert, C. H. Data-Dependent Stability of Stochastic Gradient Descent. PMLR July 2018, pp 2815–2824.

[33] Wijnhoven, R. G. J.; De With, P. H. N. Fast Training of Object Detection Using Stochastic Gradient Descent. *Proc. - Int. Conf. Pattern Recognit.* 2010, 424–427. https://doi.org/10.1109/ICPR.2010.112.

[34] Aryanmehr, S.; Karimi, M.; Boroujeni, F. Z. CVBL IRIS Gender Classification Database Image Processing and Biometric Research, Computer Vision and Biometric Laboratory (CVBL). *2018 3rd IEEE Int. Conf. Image, Vis. Comput. ICIVC 2018* 2018, 433–438. https://doi.org/10.1109/ICIVC.2018.8492757.

[35] Biza, P. Female and Male Eyes. Kaggle 2021. https://doi.org/10.34740/KAGGLE/DS/1438879.